\def\year{2019}\relax
\documentclass[letterpaper]{article} 
\usepackage{aaai19}  
\usepackage{times}  
\usepackage{helvet}  
\usepackage{courier}  
\usepackage{url}  
\usepackage{graphicx}  

\usepackage{color}
\usepackage{booktabs} 
\usepackage{makecell}
\usepackage[T1]{fontenc}
\usepackage{epstopdf}
\usepackage{multirow}
\usepackage{balance}
\usepackage{amsmath}
\usepackage{amssymb}
\usepackage{microtype}
\usepackage{enumitem}
\usepackage{wrapfig,lipsum}
\usepackage{multirow}
\usepackage{makecell}
\usepackage{wrapfig}
\usepackage[boxed, ruled,vlined,linesnumbered]{algorithm2e}
\usepackage{algpseudocode}

\usepackage{microtype}
\usepackage{bbold}
\usepackage{threeparttable}
\usepackage{bm}
\usepackage{subfigure}
\usepackage{booktabs}
\usepackage{tablefootnote}
\usepackage{tabularx}

\newcommand{\bs}[1]{\boldsymbol{#1}}
\newtheorem{thm:def}{Definition}
\newtheorem{thm:eg}{Example}
\newtheorem{thm:lem}{Lemma}
\newtheorem{thm:obs}{Observation}
\newtheorem{thm:req}{Requirement}

\newcommand{\eg}{\emph{e.g.}}

\title{Weakly-Supervised Hierarchical Text Classification}
\author{Yu Meng, Jiaming Shen, Chao Zhang, Jiawei Han\\
	University of Illinois at Urbana-Champaign, Urbana, IL, USA\\
	\{yumeng5, js2, czhang82, hanj\}@illinois.edu
}

\begin{document}

\maketitle

\begin{abstract}
	
	Hierarchical text classification, which aims to classify text documents into a given hierarchy, is an important task in many real-world applications.
	Recently, deep neural models are gaining increasing popularity for text classification due to their expressive power and minimum requirement for feature engineering. 
	However, applying deep neural networks for hierarchical text classification remains challenging, because they heavily rely on a large amount of training data and meanwhile cannot easily determine appropriate levels of documents in the hierarchical setting.  
	In this paper, we propose a weakly-supervised neural method for hierarchical text classification. 
	Our method does not require a large amount of training data but requires only easy-to-provide weak supervision signals such as a few class-related documents or keywords.
	Our method effectively leverages such weak supervision signals to generate pseudo documents for model pre-training, and then performs self-training on real unlabeled data to iteratively refine the model. 
	During the training process, our model features a hierarchical neural structure, which mimics the given hierarchy and is capable of determining the proper levels for documents with a blocking mechanism. 
	Experiments on three datasets from different domains demonstrate the efficacy of our method compared with a comprehensive set of baselines \footnote{Source code can be found at \url{https://github.com/yumeng5/WeSHClass}.}.
	
\end{abstract}


\section{Introduction}

Hierarchical text classification, which aims at classifying text documents into
classes that are organized into a hierarchy, is an important text mining and
natural language processing task. Unlike flat text classification, hierarchical
text classification considers the interrelationships among classes and allows
for organizing documents into a natural hierarchical structure. It has a wide
variety of applications such as semantic classification
\cite{Tang2015DocumentMW}, question answering \cite{Li2002LearningQC}, and web
search organization \cite{Dumais2000HierarchicalCO}.


Traditional flat text classifiers (\eg, SVM, logistic regression) have been
tailored in various ways for hierarchical text classification.  Early attempts
\cite{Ceci2006ClassifyingWD} disregard the relationships among classes and
treat hierarchical classification tasks as flat ones. Later approaches
\cite{Dumais2000HierarchicalCO,Liu2005SupportVM,Cai2004HierarchicalDC} train a
set of local classifiers and make predictions in a top-down manner, or design
global hierarchical loss functions that regularize with the hierarchy.  Most
existing efforts for hierarchical text classification rely on traditional text
classifiers. Recently, deep neural networks have demonstrated superior
performance for flat text classification. Compared with traditional
classifiers, deep neural networks
\cite{Kim2014ConvolutionalNN,Yang2016HierarchicalAN} largely reduce feature
engineering efforts by learning distributed representations that capture text
semantics.  Meanwhile, they provide stronger expressive power over traditional
classifiers, thereby yielding better performance when large amounts of training
data are available.


Motivated by the enjoyable properties of deep neural networks, we explore using
deep neural networks for hierarchical text classification. Despite the success
of deep neural models in flat text classification and their advantages over
traditional classifiers, applying them to hierarchical text classification is
nontrivial because of two major challenges. 
The first challenge is that \emph{the training data deficiency prohibits neural models from being adopted}.  
Neural models are data hungry and require humans to provide tons of carefully-labeled
documents for good performance.
In many practical scenarios, however, hand-labeling
excessive documents often requires domain expertise and can be too expensive to realize.
The second challenge is to \emph{determine the most appropriate level
	for each document in the class hierarchy}.  In hierarchical text
classification, documents do not necessarily belong to leaf nodes and may be
better assigned to intermediate nodes.  However, there are no simple ways for
existing deep neural networks to automatically determine the best granularity
for a given document. 

In this work, we propose a neural approach named \textsf{WeSHClass}, for \textbf{We}akly-\textbf{S}upervised \textbf{H}ierarchical Text
\textbf{Class}ification and address the above two challenges. 
Our approach is built upon deep neural networks, yet it
requires only a small amount of weak supervision instead of excessive training
data. Such weak supervision can be either a few (\eg, less than a dozen)
labeled documents or class-correlated keywords, which can be easily provided by
users. To leverage such weak supervision for effective classification, our
approach employs a novel pretrain-and-refine paradigm.  Specifically, in the
pre-training step, we leverage user-provided seeds to learn a spherical
distribution for each class, and then generate pseudo documents from a language
model guided by the spherical distribution. In the refinement step, we
iteratively bootstrap the global model on real unlabeled documents, which
self-learns from its own high-confident predictions.


\textsf{WeSHClass} automatically determines the most appropriate level during the
classification process by explicitly modeling the class hierarchy.
Specifically, we pre-train a local classifier at each node in the class
hierarchy, and aggregate the classifiers into a global one using self-training.
The global classifier is used to make final predictions in a top-down recursive
manner. During recursive predictions, we introduce a novel blocking mechanism,
which examines the distribution of a document over internal nodes and avoids
mandatorily pushing general documents down to leaf nodes.

Our contributions are summarized as follows:

\begin{enumerate}
	
	\item We design a method for hierarchical text classification using neural
	models under weak supervision. \textsf{WeSHClass} does not require large amounts of
	training documents but just easy-to-provide word-level or document-level
	weak supervision. In addition, it can be applied to different classification types (\eg, topics, sentiments).
	
	\item We propose a pseudo document generation module that generates
	high-quality training documents only based on weak
	supervision sources. The generated documents serve as pseudo training data
	which alleviate the training data bottleneck together with the subsequent
	self-training step.

	\item We propose a hierarchical neural model structure that mirrors the class
	taxonomy and its corresponding training method, which involves local
	classifier pre-training and global classifier self-training.  The entire
	process is tailored for hierarchical text classification, which automatically
	determines the most appropriate level of each document with a novel blocking
	mechanism.
	
	\item We conduct a thorough evaluation on three real-world datasets from
	different domains to demonstrate the effectiveness of \textsf{WeSHClass}.
	We also perform several case studies to understand the properties of
	different components in \textsf{WeSHClass}.
	
\end{enumerate}


\section{Problem Formulation}

We study hierarchical text classification that involves tree-structured class
categories. Specifically, each category can belong to at most one parent
category and can have arbitrary number of children categories.  Following the
definition in \cite{Silla2010ASO}, we consider non-mandatory leaf prediction,
wherein documents can be assigned to both internal and leaf categories in the
hierarchy.

%

Traditional supervised text classification methods rely on large amounts of
labeled documents for each class. In this work, we focus on text classification
under weak supervision.  Given a class taxonomy represented as a tree
$\mathcal{T}$, we ask the user to provide weak supervision sources (\eg, a few
class-related keywords or documents) only for each leaf class in $\mathcal{T}$.
Then we propagate the weak supervision sources upwards in $\mathcal{T}$ from
leaves to root, so that the weak supervision sources of each internal class are
an aggregation of weak supervision sources of all its descendant leaf classes.
Specifically, given $M$ leaf node classes, the supervision for each class comes
from one of the following:
\begin{enumerate}
	\item
	\emph{Word-level supervision}: $\mathcal{S} = \{S_{j}\}|_{j=1}^{M}$, where $S_{j} = \{w_{j,1}, \dots, w_{j,k}\}$ represents a set of $k$ keywords correlated with class $C_{j}$;
	\item
	\emph{Document-level supervision}: $\mathcal{D}^{L} = \{ \mathcal{D}_{j}^{L} \}|_{j=1}^{M}$, where $\mathcal{D}_{j}^{L} = \{D_{j,1}, \dots, D_{j,l} \}$ denotes a small set of $l$ ($l \ll \text{corpus size}$) labeled documents in class $C_{j}$. 
\end{enumerate}


Now we are ready to formulate the hierarchical text classification problem.
Given a text collection $\mathcal{D} = \{D_{1}, \dots, D_{N} \}$, a class
category tree $\mathcal{T}$, and weak supervisions of either $\mathcal{S}$ or
$\mathcal{D}^{L}$ for each leaf class in $\mathcal{T}$, the weakly-supervised
hierarchical text classification task aims to assign the most likely label
$C_{j} \in \mathcal{T}$ to each $D_{i} \in \mathcal{D}$, where $C_{j}$ could be
either an internal or a leaf class.

\section{Pseudo Document Generation}

To break the bottleneck of lacking abundant labeled data for model training, we
leverage user-given weak supervision to generate pseudo documents, which serve
as pseudo training data for model pre-training.  In this section, we first
introduce how to leverage weak supervision sources to model class distributions
in a spherical space, and then explain how to generate class-specific pseudo
documents based on class distributions and a language model.


\subsubsection{Modeling Class Distribution}
We model each class as a high-dimensional spherical probability distribution which has been shown effective for various tasks \cite{ZhangLLYZH017}.
We first train Skip-Gram model \cite{Mikolov2013DistributedRO} to learn
$d$-dimensional vector representations for each word in the corpus. Since
directional similarities between vectors are more effective in capturing
semantic correlations \cite{Banerjee2005ClusteringOT,Levy2015ImprovingDS}, we
normalize all the $d$-dimensional word embeddings so that they reside on a unit
sphere in $\mathbb{R}^{d}$. For each class $C_j \in \mathcal{T}$, we model the
semantics of class $C_j$ as a mixture of von Mises Fisher
(movMF) distributions \cite{Banerjee2005ClusteringOT,Gopal2014VonMC} in $\mathbb{R}^{d}$:
\begin{align*}
f(\bs{x} \mid \Theta) = \sum_{h=1}^{m} \alpha_h f_h(\bs{x} \mid \bs{\mu}_h,\kappa_h) = \sum_{h=1}^{m} \alpha_h c_d(\kappa_h)e^{\kappa_h\bs{\mu}_h^T\bs{x}},
\end{align*}where $\Theta = \{\alpha_1,\dots,\alpha_m, \bs{\mu}_1,\dots,\bs{\mu}_m, \kappa_1,\dots,\kappa_m\}$, $\forall h \in \{1,\dots,m\}$, $\kappa_h \ge 0$, $\|\bs{\mu}_h\| = 1$, and the normalization constant $c_d(\kappa_h)$ is given by 
$$
c_d(\kappa_h) = \frac{\kappa_h^{d/2-1}}{(2\pi)^{d/2} I_{d/2-1}(\kappa_h)},
$$
where $I_r(\cdot)$ represents the modified Bessel function of the first kind at
order $r$. We choose the number of components in movMF for leaf and internal
classes differently:
\begin{itemize}
	\item
	For each leaf class $C_j$, we set the number of vMF component $m=1$, and the
	resulting movMF distribution is equivalent to a single vMF distribution, whose
	two parameters, the mean direction $\bs{\mu}$ and the concentration parameter
	$\kappa$, act as semantic focus and  concentration for $C_j$.
	
	
	\item
	For each internal class $C_j$, we set the number of vMF component $m$ to be the
	number of its children classes. Recall that we only ask the user to provide
	weak supervision sources at the leaf classes, and the weak supervision source
	of $C_j$ are aggregated from its children classes. The semantics of a parent
	class can thus be seen as a mixture of the semantics of its children classes.
\end{itemize}

We first retrieve a set of keywords for each class given the weak supervision
sources, then fit movMF distributions using the embedding vectors of the
retrieved keywords. Specifically, the set of keywords are retrieved as follows:
(1) When users provide related keywords $S_j$ for each class $j$, we use the
average embedding of these seed keywords to find top-$n$  closest keywords in the
embedding space; (2) When users provide documents $\mathcal{D}^{L}_{j}$ that
are correlated with class $j$, we extract $n$ representative keywords from
$\mathcal{D}^{L}_{j}$ using tf-idf weighting.  The parameter $n$ above is set
to be the largest number that does not result in shared words across different
classes. Compared to directly using weak supervision signals, retrieving
relevant keywords for modeling class distributions has a smoothing effect which
makes our model less sensitive to the weak supervision sources. 


Let $X$ be the set of embeddings of the $n$ retrieved keywords on the unit sphere, i.e.,
$$
X = \{\bs{x}_i \in \mathbb{R}^{d} \mid \bs{x}_i \text{ drawn from }  f(\bs{x} \mid \Theta), 1\le i \le n\},
$$
we use the Expectation Maximization (EM) framework \cite{Banerjee2005ClusteringOT} to estimate the parameters $\Theta$ of the movMF distributions: 
\begin{itemize}
	\item 
	E-step:
	$$
	p(z_i = h \mid \bs{x}_i, \Theta^{(t)}) = \frac{\alpha_h^{(t)} f_h(\bs{x}_i \mid \bs{\mu}_h^{(t)},\kappa_h^{(t)})}{\sum_{h'=1}^{m} \alpha_{h'}^{(t)} f_{h'}(\bs{x}_i \mid \bs{\mu}_{h'}^{(t)},\kappa_{h'}^{(t)})},
	$$
	where $\mathcal{Z} = \{z_1,\dots,z_n\}$ is the set of hidden random variables that indicate the particular vMF distribution from which the points are sampled;
	\item 
	M-step:
	\begin{align*}
	\alpha_h^{(t+1)} &= \frac{1}{n} \sum_{i=1}^{n} p(z_i = h \mid \bs{x}_i, \Theta^{(t)}), \\
	\bs{r}_h^{(t+1)} &= \sum_{i=1}^{n} p(z_i = h \mid \bs{x}_i, \Theta^{(t)}) \bs{x}_i, \\
	\bs{\mu}_h^{(t+1)} &= \frac{\bs{r}_h^{(t+1)}}{\| \bs{r}_h^{(t+1)} \|}, \\
	\frac{I_{d/2}(\kappa_h^{(t+1)})}{I_{d/2-1}(\kappa_h^{(t+1)})} &= \frac{\| \bs{r}_h^{(t+1)} \|}{\sum_{i=1}^{n} p(z_i = h \mid \bs{x}_i, \Theta^{(t)})}.
	\end{align*}
	where we use the approximation procedure based on Newton's method \cite{Banerjee2005ClusteringOT} to derive an approximation of $\kappa_h^{(t+1)}$ because the implicit equation makes obtaining an analytic solution infeasible.
	
\end{itemize}

\subsubsection{Language Model Based Document Generation}

After obtaining the distributions for each class, we use an LSTM-based
language model \cite{Sundermeyer2012LSTMNN} to generate meaningful pseudo
documents. Specifically, we first train an LSTM language model on the entire
corpus. To generate a pseudo document of class $C_j$, we sample an embedding
vector from the movMF distribution of $C_j$ and use the closest word in
embedding space as the beginning word of the sequence. Then we feed the current
sequence to the LSTM language model to generate the next word and attach it to
the current sequence recursively \footnote{In case of long pseudo documents, we repeatedly generate several sequences and concatenate them to form the entire document.}. Since the beginning word of the pseudo
document comes directly from the class distribution, the generated
document is ensured to be correlated to $C_j$. By virtue of the mixture distribution
modeling, the semantics of every children class (if any) of $C_j$ gets a chance
to be included in the pseudo documents, so that the resulting trained
neural model will have better generalization ability.

%

%
%
%

\section{The Hierarchical Classification Model}

In this section, we introduce the hierarchical neural model and its training method under
weakly-supervised setting. 

\subsection{Local Classifier Pre-Training}
\label{sec:local}

We construct a neural classifier $M_p$ ($M_p$ could be any text classifier such
as CNNs or RNNs) for each class $C_p \in \mathcal{T}$ if $C_p$ has two or more
children classes. Intuitively, the classifier $M_p$ aims to classify the
documents assigned to $C_p$ into its children classes for more fine-grained
predictions. For each document $D_i$, the output of $M_p$ can be interpreted as
$p(D_i \in C_c \mid D_i \in C_p)$, the conditional probability of $D_i$
belonging to each children class $C_c$ of $C_p$, given $D_i$ is assigned to
$C_p$.




The local classifiers perform local text classification at internal nodes in
the hierarchy, and serve as building blocks that can be later ensembled into a
global hierarchical classifier. We generate $\beta$ pseudo documents
per class and use them to pre-train local classifiers with the
goal of providing each local classifier with a good initialization for the
subsequent self-training step. To prevent the local
classifiers from overfitting to pseudo documents and performing badly on
classifying real documents, we use pseudo labels instead of one-hot encodings
in pre-training.  Specifically, we use a hyperparameter $\alpha$ that accounts for the ``noises'' in pseudo documents, and set the pseudo label $\bs{l}_i^*$ for pseudo
document $D_i^*$ (we use $D_i^*$ instead of $D_i$ to denote a pseudo
document) as
{\small
	\begin{equation} \label{eq:1}
	l_{ij}^* = \begin{cases}
	(1-\alpha) + \alpha/m & \text{$D_i^*$ is generated from class $j$} \\
	\alpha/m & \text{otherwise}
	\end{cases}
	\end{equation}
}where $m$ is the total number of children classes at the corresponding local
classifier.  After creating pseudo labels, we pre-train each local classifier
$M_p$ of class $C_p$ using the pseudo documents for each children class of
$C_p$, by minimizing the KL divergence loss from outputs $\mathcal{Y}$ of $M_p$
to the pseudo labels $\mathcal{L}^*$, namely
$$
loss = KL(\mathcal{L}^*\|\mathcal{Y}) = \sum_i \sum_j l_{ij}^* \log \frac{l_{ij}^*}{y_{ij}}.
$$

\subsection{Global Classifier Self-Training}

\begin{figure}[h]
	\centering
	\scalebox{0.95}{
		\includegraphics[width = 0.5\textwidth]{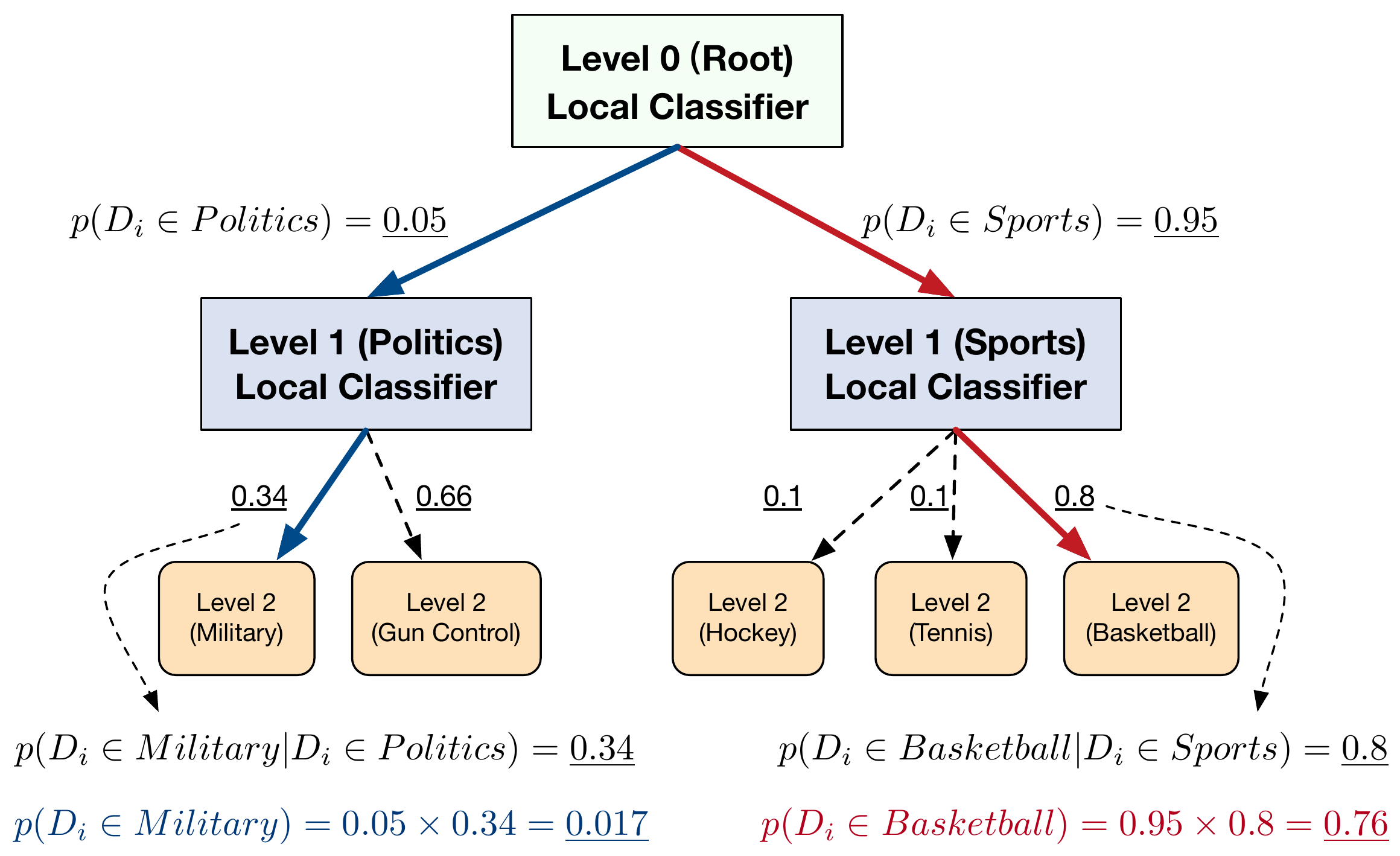}
	}
	\caption{Ensemble of local classifiers.}
	\label{fig:ensemble}
\end{figure}

At each level $k$ in the class taxonomy, we need the network to output a
probability distribution over all classes. Therefore, we construct a global
classifier $G_k$ by ensembling all local classifiers from root to level $k$.
The ensemble method is shown in Figure \ref{fig:ensemble}. The multiplication
operation conducted between parent classifier output and children classifier
output can be explained by the conditional probability formula:
\begin{align*}
p(D_i \in C_c) &= p(D_i \in C_c  \cap  D_i \in C_p) \\
&= p(D_i \in C_c \mid D_i \in C_p) p(D_i \in C_p),
\end{align*}where $D_i$ is a document; $C_c$ is one of the children classes of $C_p$. This
formula can be recursively applied so that the final prediction is the
multiplication of all local classifiers' outputs on the path from root to the
destination node.

Greedy top-down classification approaches will propagate misclassifications at
higher levels to lower levels, which can never be corrected. However, the way
we construct the global classifier assigns documents soft probability at each
level, and the final class prediction is made by jointly considering all
classifiers' outputs from root to the current level via multiplication, which
gives lower-level classifiers chances to correct misclassifications made at
higher levels. 

At each level $k$ of the class taxonomy, we first ensemble all local
classifiers from root to level $k$ to form the global classifier $G_k$, and
then use $G_k$'s prediction on all unlabeled real documents to refine itself
iteratively. Specifically, for each unlabeled document $D_i$, $G_k$ outputs a
probability distribution $y_{ij}$ of $D_i$ belonging to each class $j$ at level
$k$, and we set pseudo labels to be \cite{Xie2016UnsupervisedDE}:
\begin{equation}\label{eq:2}
l_{ij}^{**} = \frac{y_{ij}^2/f_{j}}{\sum_{j'}y_{ij'}^2/f_{j'}},
\end{equation}where $f_{j} = \sum_{i}y_{ij}$ is the soft frequency for class $j$.

The pseudo labels reflect high-confident predictions, and we use them to guide the fine-tuning of $G_k$, by iteratively (1) computing pseudo labels $\mathcal{L}^{**}$ based on $G_k$'s current predictions $\mathcal{Y}$ and (2) minimizing the KL divergence loss from $\mathcal{Y}$ to $\mathcal{L}^{**}$. This process terminates when less than $\delta \%$ of the documents in the corpus have class assignment changes.
Since $G_k$ is the ensemble of local classifiers, they are fine-tuned simultaneously via back-propagation during self-training. We will demonstrate the advantages of using global classifier over greedy approaches in the experiments.

\subsection{Blocking Mechanism}
In hierarchical classification, some documents should be
classified into internal classes because they are more related to general
topics rather than any of the more specific topics, which should be blocked at the corresponding local classifier from getting further passed to children classes.

When a document $D_i$ is  classified into an internal class $C_j$, we use the
output $\bs{q}$ of $C_j$'s local classifier to determine whether or not $D_i$
should be blocked at the current class: if $\bs{q}$ is close to a one-hot
vector, it strongly indicates that $D_i$ should be classified into the
corresponding child; if $\bs{q}$ is close to a uniform distribution, it
implies that $D_i$ is equally relevant or irrelevant to all the children
of $C_j$ and thus more likely a general document. Therefore, we
use normalized entropy as the measure for blocking. Specifically, we will block
$D_i$ from being further passed down to $C_j$'s children if
\begin{equation} \label{eq:3}
-\frac{1}{\log m}\sum_{i=1}^{m} q_i \log q_i > \gamma,
\end{equation}where $m \ge 2$ is the number of children of $C_j$; $0 \le \gamma \le 1$ is a threshold value. When $\gamma=1$, no documents will be blocked and all
documents are assigned into leaf classes.


\subsection{Inference}

The hierarchical classification model can be directly applied to classify unseen samples after training. When classifying an unseen document, the model will directly output the probability distribution of that document belonging to each class at each level in the class hierarchy. The same blocking mechanism can be applied to determine the appropriate level that the document should belong to. 

\subsection{Algorithm Summary}

Algorithm \ref{alg:modelTrain} puts the above pieces together and summarizes the overall model training process for hierarchical text classification. As shown, the overall training is proceeded in a top-down manner, from root to the final internal level. At each level, we generate pseudo documents and pseudo labels to pre-train each local classifier. Then we self-train the ensembled global classifier using its own predictions in an iterative manner.
Finally we apply blocking mechanism to block general documents, and pass the remaining documents to the next level.

\SetKwInOut{Parameter}{Parameters}

\begin{algorithm}[h]
	\small
	\caption{Overall Network Training.}
	\label{alg:modelTrain}
	\KwIn{
		A text collection $\mathcal{D} = \{D_{i}\}|_{i=1}^{N}$; a class category tree $\mathcal{T}$; weak supervisions $\mathcal{W}$ of either $\mathcal{S}$ or $\mathcal{D}^{L}$ for each leaf class in $\mathcal{T}$.
	}
	\KwOut{Class assignment $\mathcal{C} = \{(D_{i}, C_i)\}|_{i=1}^{N}$, where $C_i \in \mathcal{T}$ is the most specific class label for $D_{i}$.}
	
	Initialize $\mathcal{C} \gets \emptyset$\;
	\For{$k \gets 0$ to $max\_level-1$}  {
		$\mathcal{N} \gets$ all nodes at level $k$ of $\mathcal{T}$\;
		\ForEach{$node \in \mathcal{N}$} {
			$\mathcal{D}^{*} \gets $ Pseudo document generation\;
			$\mathcal{L}^{*} \gets$ Equation (\ref{eq:1})\;
			pre-train $node.classifier$ with $\mathcal{D}^{*}, \mathcal{L}^{*}$\;
		}
		$G_k \gets$ ensemble all classifiers from level $0$ to $k$\;
		\While{not converged}{
			$\mathcal{L}^{**} \gets$ Equation (\ref{eq:2})\;
			self-train $G_k$ with $\mathcal{D}, \mathcal{L}^{**}$\;
		}
		$\mathcal{D}_B \gets$ documents blocked based on Equation (\ref{eq:3})\;
		$\mathcal{C}_B \gets \mathcal{D}_B$'s current class assignments\;
		$\mathcal{C} \gets \mathcal{C} \cup (\mathcal{D}_B, \mathcal{C}_B)$\;
		$\mathcal{D} \gets \mathcal{D} - \mathcal{D}_B$\;
	}
	$\mathcal{C}' \gets \mathcal{D}$'s current class assignments\;
	$\mathcal{C} \gets \mathcal{C} \cup (\mathcal{D}, \mathcal{C}')$\;
	Return $\mathcal{C}$\;
\end{algorithm}

\section{Experiments}

\subsection{Experiment Settings}

\subsubsection{Datasets and Evaluation Metrics}
We use three corpora from three different domains to evaluate the performance of our proposed method:
\begin{itemize}
	\item \textbf{The New York Times (NYT):} We crawl $13,081$ news articles using the
	New York Times API \footnote{\url{http://developer.nytimes.com/}}. This news
	corpus covers $5$ super-categories and $25$ sub-categories. 
	\item \textbf{arXiv:} We crawl paper abstracts from arXiv
	website\footnote{\url{https://arxiv.org/}} and keep all abstracts that belong
	to only one category. Then we include all sub-categories with more than
	$1,000$ documents out of $3$ largest super-categories and end up with
	$230,105$ abstracts from $53$ sub-categories.
	\item \textbf{Yelp Review:}
	We use the Yelp Review Full dataset \cite{Zhang2015CharacterlevelCN} and take
	its testing portion as our dataset. The dataset contains $50,000$ documents
	evenly distributed into $5$ sub-categories, corresponding to user ratings from
	$1$ star to $5$ stars. We consider $1$ and $2$ stars as ``negative'', $3$ stars
	as ``neutral'', $4$ and $5$ stars as ``positive'', so we end up with $3$
	super-categories.
\end{itemize}

Table \ref{tab:dataset} provides the statistics of the three datasets; Table
\ref{tab:nyt_sub} and \ref{tab:arxiv_sub} show some sample sub-categories of
\textbf{NYT} and \textbf{arXiv} datasets. We use Micro-F1 and Macro-F1 scores as metrics for classification performances.

\begin{table}[!t]
	\caption{Dataset Statistics.}
	\label{tab:dataset}
	\scalebox{0.80}{
		\begin{tabular}{*{4}{c}}
			\toprule
			Corpus name & \makecell{\# classes \\(level $1$ + level $2$)} & \# docs & Avg. doc length \\
			\midrule
			NYT & $5+25$ & $13,081$ & $778$ \\
			arXiv & $3+53$ & $230,105$ & $129$ \\
			Yelp Review & $3+5$ & $50,000$ & $157$ \\
			\bottomrule
		\end{tabular}
	}
\end{table}

\begin{table}[!h]
	\caption{Sample subcategories of NYT Dataset.}
	\label{tab:nyt_sub}
	\scalebox{0.80}{
		\begin{tabular}{cc}
			\toprule
			\makecell{Super-category (\# children)} & Sub-category \\
			\midrule
			Politics ($9$) & abortion, surveillance, immigration, \dots\\
			Arts ($4$) & dance, television, music, movies \\
			Business ($4$) & stocks, energy companies, economy, \dots \\
			Science ($2$) & cosmos, environment \\
			Sports ($7$) & hockey, basketball, tennis, golf, \dots \\
			\bottomrule
		\end{tabular}
	}
\end{table}

\begin{table}[!h]
	\caption{Sample subcategories of arXiv Dataset.}
	\label{tab:arxiv_sub}
	\scalebox{0.80}{
		\begin{tabular}{cc}
			\toprule
			\makecell{Super-category (\# children)} & Sub-category \\
			\midrule
			Math ($25$) & math.NA, math.AG, math.FA, \dots\\
			Physics ($10$) & physics.optics, physics.flu-dyn, \dots \\
			CS ($18$) & cs.CV, cs.GT, cs.IT, cs.AI, cs.DC, \dots \\
			\bottomrule
		\end{tabular}
	}
\end{table}



\subsubsection{Baselines}
We compare our proposed method with a wide range of baseline models, described
as below:

\begin{itemize}
	\item
	\textbf{Hier-Dataless} \cite{Song2014OnDH}: Dataless hierarchical text
	classification
	\footnote{\url{https://github.com/CogComp/cogcomp-nlp/tree/master/dataless-classifier}}
	can only take \textbf{word-level} supervision sources. It embeds
	both class labels and documents in a semantic space using Explicit Semantic
	Analysis \cite{Gabrilovich2007ComputingSR} on Wikipedia articles,
	and assigns the nearest label to each document in
	the semantic space. We try both the top-down approach and bottom-up approach,
	with and without the bootstrapping procedure, and finally report the best
	performance.
	\item
	\textbf{Hier-SVM} \cite{Dumais2000HierarchicalCO,Liu2005SupportVM}:
	Hierarchical SVM can only take \textbf{document-level} supervision
	sources. It decomposes the training tasks according to the class taxonomy,
	where each local SVM is trained to distinguish sibling categories that
	share the same parent node.
	\item 
	\textbf{CNN} \cite{Kim2014ConvolutionalNN}: The CNN text classification model
	\footnote{\url{https://github.com/alexander-rakhlin/CNN-for-Sentence-Classification-in-Keras}}
	can only take \textbf{document-level} supervision sources. 
	\item
	\textbf{WeSTClass} \cite{Meng2018WeaklySN}: Weakly-supervised neural text classification can take both \textbf{word-level} and \textbf{document-level} supervision sources. It first generates bag-of-words pseudo documents for neural model pre-training, then bootstraps the model on unlabeled data.
	\item 
	\textbf{No-global}: This is a variant of \textsf{WeSHClass} without the global
	classifier, i.e., each document is pushed down with local classifiers in a
	greedy manner.
	\item 
	\textbf{No-vMF}: This is a variant of \textsf{WeSHClass} without using movMF
	distribution to model class semantics, i.e., we randomly select one word from
	the keyword set of each class as the beginning word when generating pseudo
	documents.
	
	\item 
	\textbf{No-selftrain}: This is a variant of \textsf{WeSHClass} without self-training module, i.e., after pre-training each local classifier, we directly ensemble them as a global classifier at each level to classify unlabeled documents.
\end{itemize}

\subsubsection{Parameter Settings}
For all datasets, we use Skip-Gram model \cite{Mikolov2013DistributedRO} to
train $100$-dimensional word embeddings for both movMF distributions modeling
and classifier input embeddings. We set the pseudo label parameter $\alpha
= 0.2$, the number of pseudo documents per class for pre-training $\beta =
500$, and the self-training stopping criterion $\delta = 0.1$. We set the
blocking threshold $\gamma = 0.9$ for \textbf{NYT} dataset where general documents exist and
$\gamma = 1$ for the other two.

Although our proposed method can use any neural model as local classifiers, we
empirically find that CNN model always results in better performances than RNN
models, such as LSTM \cite{Hochreiter1997LongSM} and Hierarchical Attention
Networks \cite{Yang2016HierarchicalAN}. Therefore, we report the performance of
our method by using CNN model with one convolutional layer as local
classifiers. Specifically, the filter window sizes are $2,3,4,5$ with $20$
feature maps each. Both the pre-training and the self-training steps are
performed using SGD with batch size $256$.

\subsubsection{Weak Supervision Settings}
The seed information we use as weak supervision for different datasets are
described as follows: (1) When the supervision source is \textbf{class-related
	keywords}, we select $3$ keywords for each leaf class; (2) When the
supervision source is \textbf{labeled documents}, we randomly sample $c$
documents of each leaf class from the corpus ($c = 3$ for \textbf{NYT} and \textbf{arXiv}; $c = 10$ for \textbf{Yelp Review}) and use them as
given labeled documents. To alleviate the randomness, we repeat the document
selection process $10$ times and show the performances with average and
standard deviation values. 

We list the keyword supervisions of some sample classes for \textbf{NYT} dataset as follows: \textbf{Immigration} (immigrants, immigration, citizenship); \textbf{Dance} (ballet, dancers, dancer); \textbf{Environment} (climate, wildlife, fish).

\subsection{Quantitative Comparision}
We show the overall text classification results in Table \ref{tab:all}. \textsf{WeSHClass} achieves the overall best performance among all the baselines on the three datasets. Notably, when the supervision source is \textbf{class-related
keywords}, \textsf{WeSHClass} outperforms \textbf{Hier-Dataless} and \textbf{WeSTClass}, which shows that \textsf{WeSHClass} can better leverage word-level supervision sources in hierarchical text classification. When the supervision source is \textbf{labeled documents}, \textsf{WeSHClass} has not only higher average performance, but also better stability than the supervised baselines. This demonstrates that when training documents are extremely limited, \textsf{WeSHClass} can better leverage the insufficient supervision for good performances and is less sensitive to seed documents.

Comparing \textsf{WeSHClass} with several ablations, \textbf{No-global},
\textbf{No-vMF} and \textbf{No-self-train}, we observe the effectiveness of
the following components: (1) ensemble of local classifiers, (2) modeling class
semantics as movMF distributions, and (3) self-training. The results
demonstrate that all these components contribute to the performance of \textsf{WeSHClass}.

\begin{table*}[!t]
	\centering
	\caption{Macro-F1 and Micro-F1 scores for all methods on three datasets, under two types of weak supervisions.}
	\label{tab:all}
	\scalebox{0.68}{
		\begin{tabular}{*{13}{c}}
			\toprule
			\textbf{Methods} & \multicolumn{4}{c}{\textbf{NYT}} & \multicolumn{4}{c}{\textbf{arXiv}} & \multicolumn{4}{c}{\textbf{Yelp Review}}\\
			\cmidrule(l){2-13}
			& \multicolumn{2}{c}{\textbf{KEYWORDS}} & \multicolumn{2}{c}{\textbf{DOCS}} & \multicolumn{2}{c}{\textbf{KEYWORDS}} & \multicolumn{2}{c}{\textbf{DOCS}} & \multicolumn{2}{c}{\textbf{KEYWORDS}} & \multicolumn{2}{c}{\textbf{DOCS}} \\
			\cmidrule(l){2-13}
			& Macro & Micro & \makecell{Macro\\Avg. (Std.)} & \makecell{Micro\\Avg. (Std.)} & Macro & Micro & \makecell{Macro\\Avg. (Std.)} & \makecell{Micro\\Avg. (Std.)} & Macro & Micro & \makecell{Macro\\Avg. (Std.)} & \makecell{Micro\\Avg. (Std.)} \\
			\midrule
			
			Hier-Dataless & $0.593$ & $0.811$ & - & - & $0.374$ & $0.594$ & - & - & $0.284$ & $0.312$ & - & - \\
			
			Hier-SVM & - & - & $0.142\ (0.016)$ & $0.469\ (\mathbf{0.012})$ & - & - & $0.049\ (\mathbf{0.001})$ & $0.443\ (\mathbf{0.006})$ & - & - & $0.220\ (0.082)$ & $0.310\ (0.113)$ \\
			
			CNN & - & - & $0.165\ (0.027)$ & $0.329\ (0.097)$ & - & - & $0.124\ (0.014)$ & $0.456\ (0.023)$ & - & - & $0.306\ (0.028)$ & $0.372\ (0.028)$ \\ 
			
			WeSTClass & $0.386$ & $0.772$ & $0.479\ (0.027)$ & $0.728\ (0.036)$ & $0.412$ & $0.642$ & $0.264\ (0.016)$ & $0.547\ (0.009)$ & $0.348$ & $0.389$ & $0.345\ (0.027)$ & $ 0.388\ (0.033)$\\
			
			No-global & $0.618$ & $0.843$ & $0.520\ (0.065)$ & $0.768\ (0.100)$ & $0.442$ & $0.673$ & $0.264\ (0.020)$ & $0.581\ (0.017)$ & $0.391$ & $0.424$ & $0.369\ (0.022)$  & $0.403\ (0.016)$\\
			
			No-vMF & $0.628$ & $0.862$ & $0.527\ (0.031)$ & $0.825\ (0.032)$ & $0.406$ & $0.665$ & $0.255\ (0.015)$ & $0.564\ (0.012)$ & $0.410$ & $0.457$ & $0.372\ (0.029)$ & $0.407\ (0.015)$\\
			
			No-self-train & $0.550$ & $0.787$ & $0.491\ (0.036)$ & $0.769\ (0.039)$ & $0.395$ & $0.635$ & $0.234\ (0.013)$ & $0.535\ (0.010)$ & $0.362$ & $0.408$ & $0.348\ (0.030)$ & $0.382\ (0.022)$\\
			
			\midrule
			
			\textsf{WeSHClass} & {$\mathbf{0.632}$} & {$\mathbf{0.874}$} & {$\mathbf{0.532}\ (\mathbf{0.015})$} &  {$\mathbf{0.827}\ (\mathbf{0.012})$} & {$\mathbf{0.452}$} & {$\mathbf{0.692}$} & {$\mathbf{0.279}\ (0.010)$} & {$\mathbf{0.585}\ (0.009)$} & {$\mathbf{0.423}$} & {$\mathbf{0.461}$} & {$\mathbf{0.375}\ (\mathbf{0.021})$} & {$\mathbf{0.410}\ (\mathbf{0.014})$}\\
			
			\bottomrule
			
		\end{tabular}
	}
\end{table*}

\subsection{Component-Wise Evaluation}

In this subsection, we conduct a series of breakdown experiments on \textbf{NYT} dataset using \textbf{class-related keywords} as weak supervision to further
investigate different components in our proposed method. We obtain similar results on the other two datasets.

\subsubsection{Pseudo Documents Generation}
The quality of the generated pseudo documents is critical to our model, since
high-quality pseudo documents provide a good model initialization. Therefore,
we are interested in which pseudo document generation method gives our model
best initialization for the subsequent self-training step. We compare our 
document generation strategy (movMF + LSTM language model) with the following two
methods:
\begin{itemize}
	\item
	Bag-of-words \cite{Meng2018WeaklySN}: The pseudo documents are generated from a mixture of background unigram distribution and class-related keywords distribution.
	\item
	Bag-of-words + reordering: We first generate bag-of-words pseudo documents as in the previous method, and then use the globally trained LSTM language model to reorder the pseudo documents by greedily putting the word with the highest probability at the end of the current sequence. The beginning word is randomly chosen.
\end{itemize}

We showcase some generated pseudo document snippets of class ``politics''
for \textbf{NYT} dataset using different methods in Table
\ref{tab:pseudo}.  Bag-of-words method generates pseudo documents without word order information; bag-of-words method with reordering generates
text of high quality at the beginning, but poor near the end, which is probably
because the ``proper'' words have been used at the beginning, but the remaining
words are crowded at the end implausibly; our method generates text of high
quality.

To compare the generalization ability of the pre-trained models with different
pseudo documents, we show their subsequent self-training process (at level $1$)
in Figure \ref{fig:pseudo}. We notice that our strategy not only
makes self-training converge faster, but also has better final performance. 

\begin{table*}
	\centering
	\caption{Sample generated pseudo document snippets of class ``politics'' for \textbf{NYT} dataset.}
	\label{tab:pseudo}
	\scalebox{0.83}{
		\begin{tabular}{c|cc|cc|cc}
			\toprule
			Doc \# & \multicolumn{2}{c|}{Bag-of-words} & \multicolumn{2}{c|}{Bag-of-words + reordering} &  \multicolumn{2}{c}{movMF + LSTM language model}\\
			\midrule
			1 & \multicolumn{2}{p{6cm}|}{he's cup abortion bars have pointed use of lawsuits involving smoothen bettors rights in the federal exchange, limewire \dots} 
			& \multicolumn{2}{p{6cm}|}{the clinicians pianists said that the legalizing of the profiling of the \dots abortion abortion abortion identification abortions \dots} 
			& \multicolumn{2}{p{6cm}}{abortion rights is often overlooked by the president's 30-feb format of a moonjock period that offered him the rules to \dots} \\
			\midrule
			2 & \multicolumn{2}{p{6cm}|}{first tried to launch the agent in immigrants were in a lazar and lakshmi definition of yerxa riding this we get very coveted as \dots} 
			&
			\multicolumn{2}{p{6cm}|}{majorities and clintons legalization, moderates and tribes lawfully \dots lawmakers clinics immigrants immigrants immigrants \dots} 
			&\multicolumn{2}{p{6cm}}{immigrants who had been headed to the united states in benghazi, libya, saying that mr. he making comments describing \dots} 
			\\
			\midrule
			3 & \multicolumn{2}{p{6cm}|}{the september crew members budget security administrator lat coequal representing a federal customer, identified the bladed \dots}  
			& \multicolumn{2}{p{6cm}|}{the impasse of allowances overruns pensions entitlement \dots funding financing budgets budgets budgets budgets taxpayers \dots}  
			& \multicolumn{2}{p{6cm}}{budget increases on oil supplies have grown more than a ezio of its 20 percent of energy spaces, producing plans by 1 billion \dots}
			\\
			\bottomrule
		\end{tabular}
	}
\end{table*}

\subsubsection{Global Classifier and Self-training}
We proceed to study why using self-trained global classifier on the ensemble of local classifiers is better than
greedy approach. We show the self-training
procedure of the global classifier at the final level in Figure \ref{fig:self}, where we
demonstrate the classification accuracy at level $1$ (super-categories), level
$2$ (sub-categories) and of all classes.  Since at the final level, all local
classifiers are ensembled to construct the global classifier, self-training of
the global classifier is the joint training of all local
classifiers. The result shows that the ensemble of local classifiers for joint
training is beneficial for improving the accuracy at all levels. If a greedy
approach is used, however, higher-level classifiers will not be updated during
lower-level classification, and misclassification at higher levels cannot
be corrected.


\subsubsection{Blocking During Self-training}

\begin{figure}[h]
	\subfigure[Pseudo documents generation][\scriptsize{Pseudo documents generation}]{
		\label{fig:pseudo}
		\includegraphics[width = 0.22\textwidth]{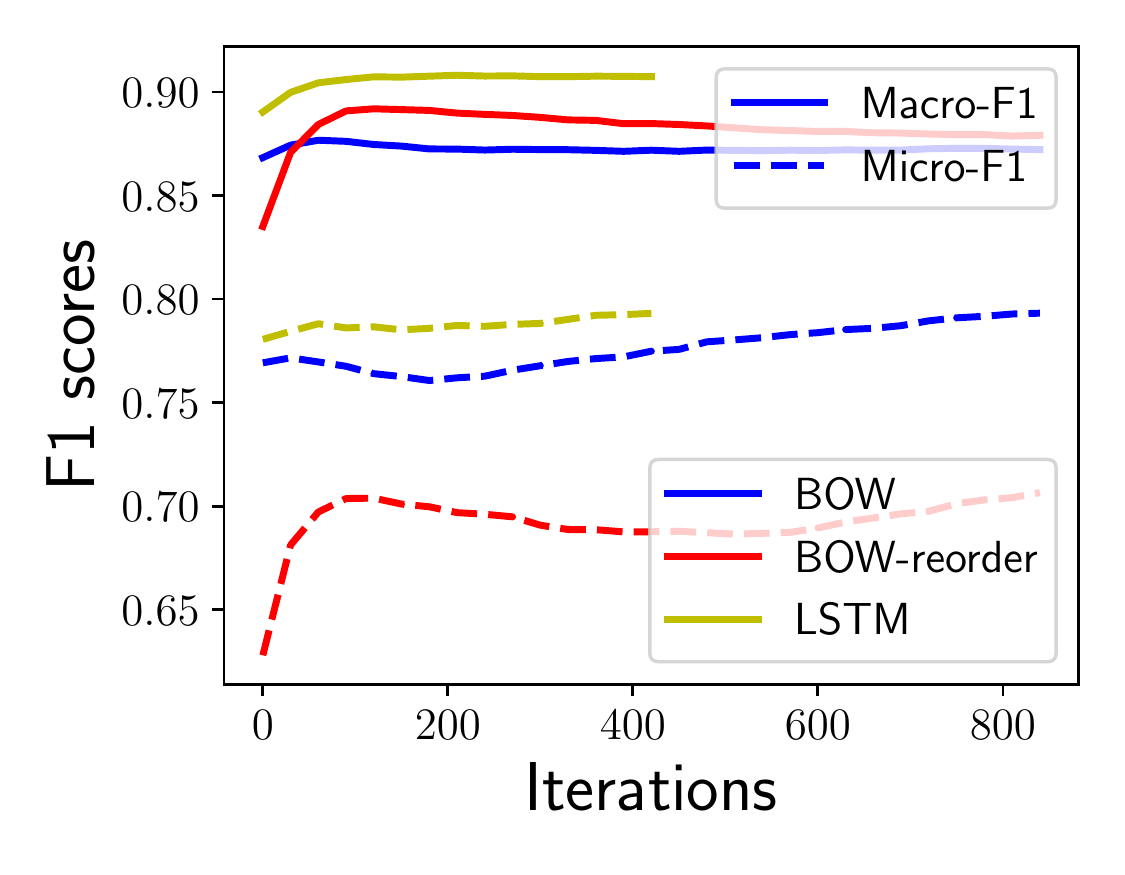}
	}
	\subfigure[Global classifier self-training][\scriptsize{Global classifier self-training}]{
		\label{fig:self}
		\includegraphics[width = 0.22\textwidth]{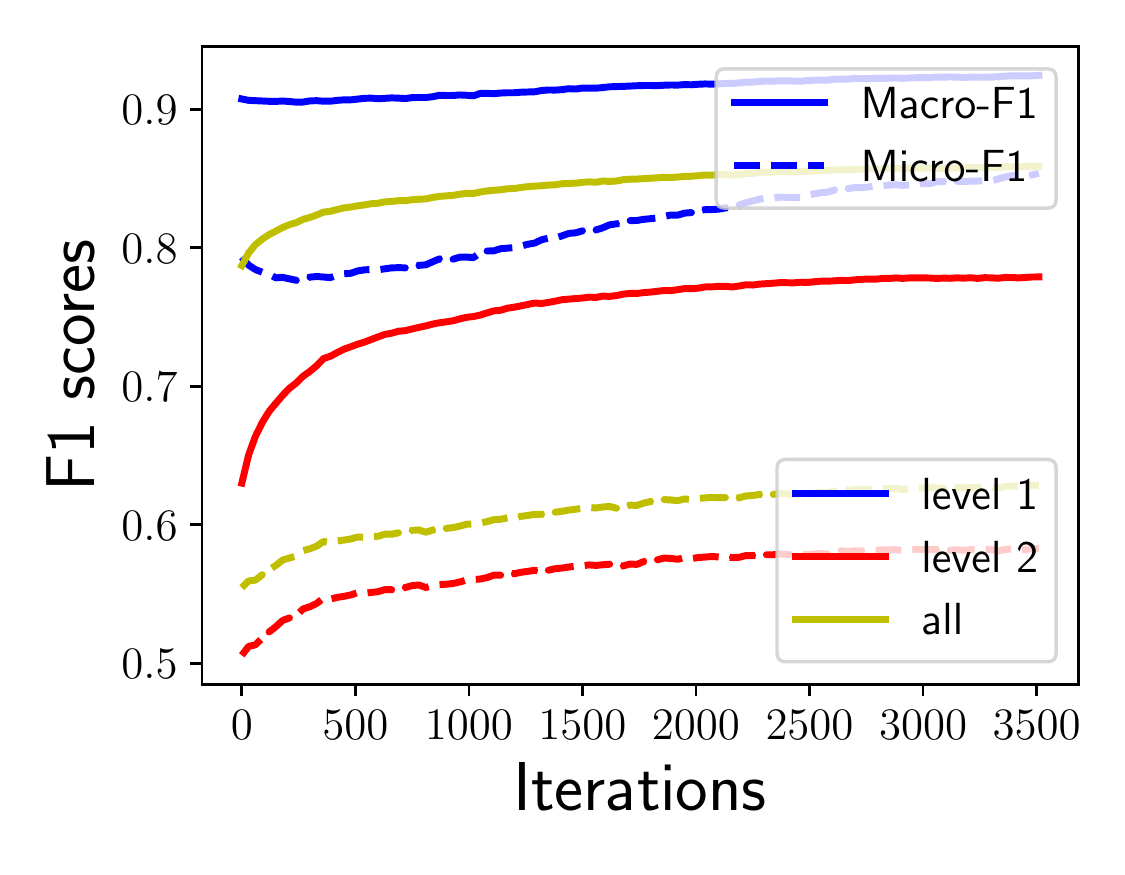}
	}
	\subfigure[Average normalized entropy][\scriptsize{Average normalized entropy}]{
		\label{fig:ent}
		\includegraphics[width = 0.22\textwidth]{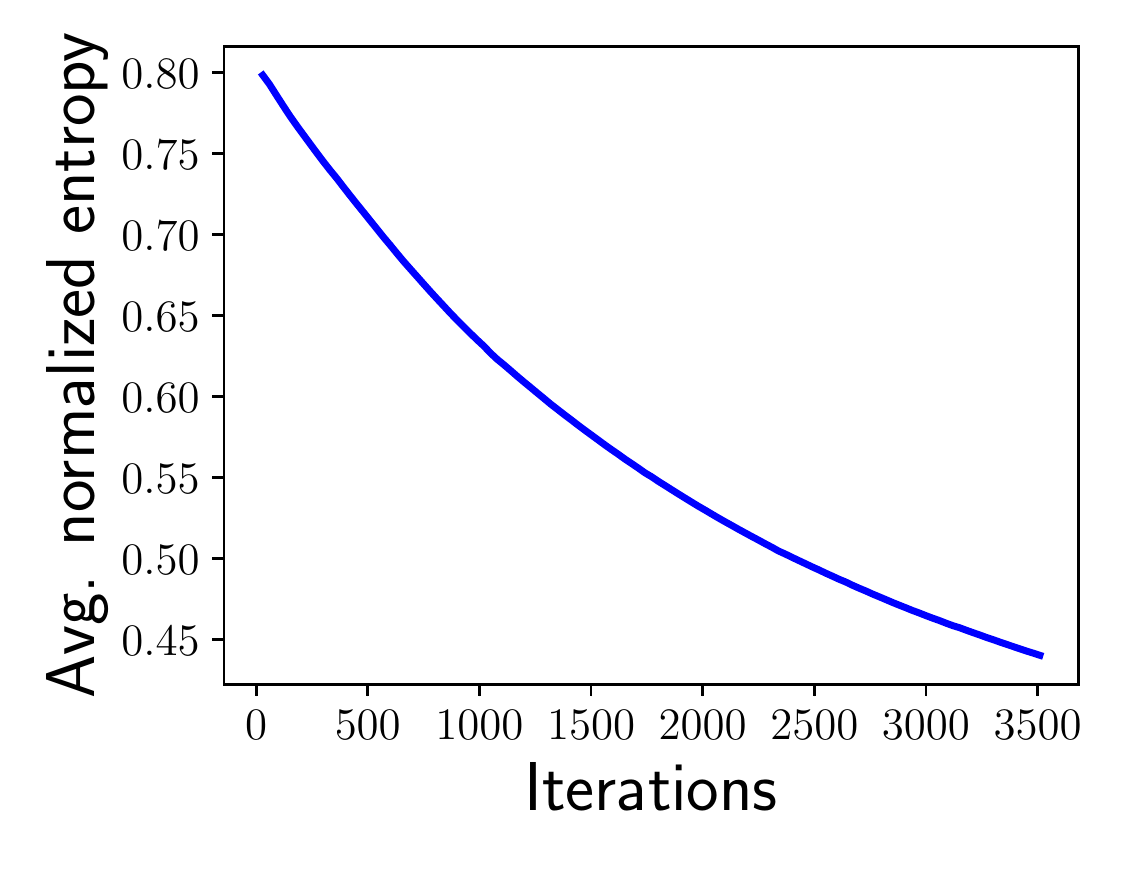}
	}
	\subfigure[Number of blocked documents][\scriptsize{Number of blocked documents}]{
		\label{fig:block}
		\includegraphics[width = 0.22\textwidth]{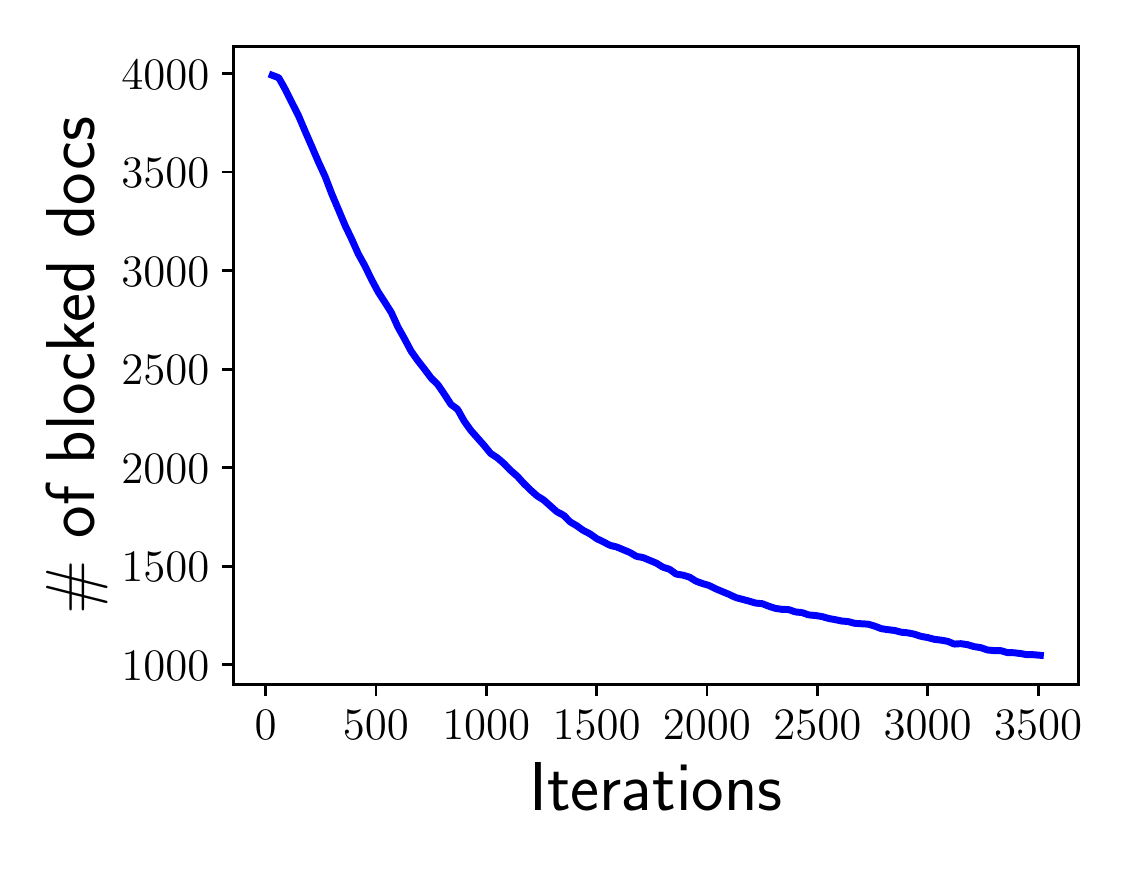}
	}
	\caption{Component-wise evaluation on \textbf{NYT} dataset.}
	\label{fig:component}
\end{figure}

We demonstrate the dynamics of the blocking mechanism during self-training. Figure
\ref{fig:ent} shows the average normalized entropy of the corresponding
local classifier output for each document in  \textbf{NYT}
dataset, and Figure \ref{fig:block} shows the total number of blocked documents during the self-training procedure at the final level. Recall that we enhance high-confident
predictions to refine our model during self-training. Therefore, the average
normalized entropy decreases during self-training, implying there is less
uncertainty in the outputs of our model. Correspondingly, fewer documents will
be blocked, resulting in more available documents for self-training.


\section{Related Work}

\subsection{Weakly-Supervised Text Classification} 
There exist some previous studies that use either word-based supervision or limited amount of labeled
documents as weak supervision sources for the text classification task. 
WeSTClass \cite{Meng2018WeaklySN} leverages both types of supervision
sources. It applies a similar procedure of pre-training the network with
pseudo documents followed by self-training on unlabeled data. 
Descriptive LDA \cite{Chen2015DatalessTC} applies an LDA model to infer Dirichlet priors from given keywords as category descriptions. 
The Dirichlet priors guide LDA to induce the category-aware
topics from unlabeled documents for classification.
\cite{Ganchev2010PosteriorRF} propose to encode prior knowledge and indirect
supervision in constraints on posteriors of latent variable probabilistic
models. Predictive text embedding \cite{Tang2015PTEPT} utilizes both labeled
and unlabeled documents to learn text embedding specifically for a task.
Labeled data and word co-occurrence information are first represented as a
large-scale heterogeneous text network and then embedded into a low dimensional
space. The learned embedding are fed to logistic regression classifiers for
classification. None of the above methods are specifically designed for hierarchical classification. 

\subsection{Hierarchical Text Classification}

There have been efforts on using SVM for hierarchical classification.
\cite{Dumais2000HierarchicalCO,Liu2005SupportVM} propose to use local SVMs that
are trained to  distinguish the children classes of the same parent node so
that the hierarchical classification task is decomposed into several flat
classification tasks. \cite{Cai2004HierarchicalDC} define hierarchical loss
function and apply cost-sensitive learning to generalize SVM learning for
hierarchical classification. A graph-CNN based deep learning model is proposed
in \cite{Peng2018LargeScaleHT} to convert text to graph-of-words, on which the
graph convolution operations are applied for feature extraction. FastXML
\cite{Prabhu2014FastXMLAF} is designed for extremely large label space. It
learns a hierarchy of training instances and optimizes a ranking-based
objective at each node of the hierarchy.  The above methods rely heavily on the
quantity and quality of training data for good performance, while \textsf{WeSHClass}
does not require much training data but only weak supervision from users.

Hierarchical dataless classification \cite{Song2014OnDH} uses class-related
keywords as class descriptions, and projects classes and documents into the
same semantic space by retrieving Wikipedia concepts.  Classification can be
performed in both top-down and bottom-up manners, by measuring the vector
similarity between documents and classes. Although hierarchical dataless
classification does not rely on massive training data as well, its performance
is highly influenced by the text similarity between the distant supervision
source (Wikipedia) and the given unlabeled corpus.

\section{Conclusions}

We proposed a weakly-supervised hierarchical text classification method \textsf{WeSHClass}. Our designed hierarchical network structure and training method can effectively leverage (1) different types of weak supervision sources to generate
high-quality pseudo documents for better model generalization ability, and (2)
class taxonomy for better performances than flat methods and greedy approaches.
\textsf{WeSHClass} outperforms various supervised and weakly-supervised baselines in
three datasets from different domains, which demonstrates the practical value
of \textsf{WeSHClass} in real-world applications.  In the future, it is interesting to
study what kinds of weak supervision are most effective for the hierarchical
text classification task and how to combine multiple sources together to
achieve even better performance.

\section*{Acknowledgements}
This research is sponsored in part by U.S. Army Research Lab. under Cooperative Agreement No. W911NF-09-2-0053 (NSCTA), DARPA under Agreement No. W911NF-17-C-0099, National Science Foundation IIS 16-18481, IIS 17-04532, and IIS-17-41317, DTRA HDTRA11810026, and grant 1U54GM114838 awarded by NIGMS through funds provided by the trans-NIH Big Data to Knowledge (BD2K) initiative (www.bd2k.nih.gov). We thank anonymous reviewers for valuable and insightful feedback.

\bibliographystyle{aaai}
\bibliography{ref}

\begin{thebibliography}{}

\bibitem[\protect\citeauthoryear{Banerjee \bgroup et al\mbox.\egroup
  }{2005}]{Banerjee2005ClusteringOT}
Banerjee, A.; Dhillon, I.~S.; Ghosh, J.; and Sra, S.
\newblock 2005.
\newblock Clustering on the unit hypersphere using von mises-fisher
  distributions.
\newblock {\em Journal of Machine Learning Research} 6:1345--1382.

\bibitem[\protect\citeauthoryear{Cai and Hofmann}{2004}]{Cai2004HierarchicalDC}
Cai, L., and Hofmann, T.
\newblock 2004.
\newblock Hierarchical document categorization with support vector machines.
\newblock In {\em CIKM}.

\bibitem[\protect\citeauthoryear{Ceci and
  Malerba}{2006}]{Ceci2006ClassifyingWD}
Ceci, M., and Malerba, D.
\newblock 2006.
\newblock Classifying web documents in a hierarchy of categories: a
  comprehensive study.
\newblock {\em Journal of Intelligent Information Systems} 28:37--78.

\bibitem[\protect\citeauthoryear{Chen \bgroup et al\mbox.\egroup
  }{2015}]{Chen2015DatalessTC}
Chen, X.; Xia, Y.; Jin, P.; and Carroll, J.~A.
\newblock 2015.
\newblock Dataless text classification with descriptive lda.
\newblock In {\em AAAI}.

\bibitem[\protect\citeauthoryear{Dumais and
  Chen}{2000}]{Dumais2000HierarchicalCO}
Dumais, S.~T., and Chen, H.
\newblock 2000.
\newblock Hierarchical classification of web content.
\newblock In {\em SIGIR}.

\bibitem[\protect\citeauthoryear{Gabrilovich and
  Markovitch}{2007}]{Gabrilovich2007ComputingSR}
Gabrilovich, E., and Markovitch, S.
\newblock 2007.
\newblock Computing semantic relatedness using wikipedia-based explicit
  semantic analysis.
\newblock In {\em IJCAI}.

\bibitem[\protect\citeauthoryear{Ganchev \bgroup et al\mbox.\egroup
  }{2010}]{Ganchev2010PosteriorRF}
Ganchev, K.; Graça, J.; Gillenwater, J.; and Taskar, B.
\newblock 2010.
\newblock Posterior regularization for structured latent variable models.
\newblock {\em Journal of Machine Learning Research} 11:2001--2049.

\bibitem[\protect\citeauthoryear{Gopal and Yang}{2014}]{Gopal2014VonMC}
Gopal, S., and Yang, Y.
\newblock 2014.
\newblock Von mises-fisher clustering models.
\newblock In {\em ICML}.

\bibitem[\protect\citeauthoryear{Hochreiter and
  Schmidhuber}{1997}]{Hochreiter1997LongSM}
Hochreiter, S., and Schmidhuber, J.
\newblock 1997.
\newblock Long short-term memory.
\newblock {\em Neural Computation} 9:1735--1780.

\bibitem[\protect\citeauthoryear{Kim}{2014}]{Kim2014ConvolutionalNN}
Kim, Y.
\newblock 2014.
\newblock Convolutional neural networks for sentence classification.
\newblock In {\em EMNLP}.

\bibitem[\protect\citeauthoryear{Levy, Goldberg, and
  Dagan}{2015}]{Levy2015ImprovingDS}
Levy, O.; Goldberg, Y.; and Dagan, I.
\newblock 2015.
\newblock Improving distributional similarity with lessons learned from word
  embeddings.
\newblock {\em TACL} 3:211--225.

\bibitem[\protect\citeauthoryear{Li and Roth}{2002}]{Li2002LearningQC}
Li, X., and Roth, D.
\newblock 2002.
\newblock Learning question classifiers.
\newblock In {\em COLING}.

\bibitem[\protect\citeauthoryear{Liu \bgroup et al\mbox.\egroup
  }{2005}]{Liu2005SupportVM}
Liu, T.-Y.; Yang, Y.; Wan, H.; Zeng, H.-J.; Chen, Z.; and Ma, W.-Y.
\newblock 2005.
\newblock Support vector machines classification with a very large-scale
  taxonomy.
\newblock {\em SIGKDD Explorations} 7:36--43.

\bibitem[\protect\citeauthoryear{Meng \bgroup et al\mbox.\egroup
  }{2018}]{Meng2018WeaklySN}
Meng, Y.; Shen, J.; Zhang, C.; and Han, J.
\newblock 2018.
\newblock Weakly-supervised neural text classification.
\newblock In {\em CIKM}.

\bibitem[\protect\citeauthoryear{Mikolov \bgroup et al\mbox.\egroup
  }{2013}]{Mikolov2013DistributedRO}
Mikolov, T.; Sutskever, I.; Chen, K.; Corrado, G.~S.; and Dean, J.
\newblock 2013.
\newblock Distributed representations of words and phrases and their
  compositionality.
\newblock In {\em NIPS}.

\bibitem[\protect\citeauthoryear{Peng \bgroup et al\mbox.\egroup
  }{2018}]{Peng2018LargeScaleHT}
Peng, H.; Li, J.; He, Y.; Liu, Y.; Bao, M.; Wang, L.; Song, Y.; and Yang, Q.
\newblock 2018.
\newblock Large-scale hierarchical text classification with recursively
  regularized deep graph-cnn.
\newblock In {\em WWW}.

\bibitem[\protect\citeauthoryear{Prabhu and Varma}{2014}]{Prabhu2014FastXMLAF}
Prabhu, Y., and Varma, M.
\newblock 2014.
\newblock Fastxml: a fast, accurate and stable tree-classifier for extreme
  multi-label learning.
\newblock In {\em KDD}.

\bibitem[\protect\citeauthoryear{Silla and Freitas}{2010}]{Silla2010ASO}
Silla, C.~N., and Freitas, A.~A.
\newblock 2010.
\newblock A survey of hierarchical classification across different application
  domains.
\newblock {\em Data Mining and Knowledge Discovery} 22:31--72.

\bibitem[\protect\citeauthoryear{Song and Roth}{2014}]{Song2014OnDH}
Song, Y., and Roth, D.
\newblock 2014.
\newblock On dataless hierarchical text classification.
\newblock In {\em AAAI}.

\bibitem[\protect\citeauthoryear{Sundermeyer, Schl{\"u}ter, and
  Ney}{2012}]{Sundermeyer2012LSTMNN}
Sundermeyer, M.; Schl{\"u}ter, R.; and Ney, H.
\newblock 2012.
\newblock Lstm neural networks for language modeling.
\newblock In {\em INTERSPEECH}.

\bibitem[\protect\citeauthoryear{Tang, Qin, and Liu}{2015}]{Tang2015DocumentMW}
Tang, D.; Qin, B.; and Liu, T.
\newblock 2015.
\newblock Document modeling with gated recurrent neural network for sentiment
  classification.
\newblock In {\em EMNLP}.

\bibitem[\protect\citeauthoryear{Tang, Qu, and Mei}{2015}]{Tang2015PTEPT}
Tang, J.; Qu, M.; and Mei, Q.
\newblock 2015.
\newblock Pte: Predictive text embedding through large-scale heterogeneous text
  networks.
\newblock In {\em KDD}.

\bibitem[\protect\citeauthoryear{Xie, Girshick, and
  Farhadi}{2016}]{Xie2016UnsupervisedDE}
Xie, J.; Girshick, R.~B.; and Farhadi, A.
\newblock 2016.
\newblock Unsupervised deep embedding for clustering analysis.
\newblock In {\em ICML}.

\bibitem[\protect\citeauthoryear{Yang \bgroup et al\mbox.\egroup
  }{2016}]{Yang2016HierarchicalAN}
Yang, Z.; Yang, D.; Dyer, C.; He, X.; Smola, A.~J.; and Hovy, E.~H.
\newblock 2016.
\newblock Hierarchical attention networks for document classification.
\newblock In {\em HLT-NAACL}.

\bibitem[\protect\citeauthoryear{Zhang \bgroup et al\mbox.\egroup
  }{2017}]{ZhangLLYZH017}
Zhang, C.; Liu, L.; Lei, D.; Yuan, Q.; Zhuang, H.; Hanratty, T.; and Han, J.
\newblock 2017.
\newblock Triovecevent: Embedding-based online local event detection in
  geo-tagged tweet streams.
\newblock In {\em KDD}.

\bibitem[\protect\citeauthoryear{Zhang, Zhao, and
  LeCun}{2015}]{Zhang2015CharacterlevelCN}
Zhang, X.; Zhao, J.~J.; and LeCun, Y.
\newblock 2015.
\newblock Character-level convolutional networks for text classification.
\newblock In {\em NIPS}.

\end{thebibliography}

\end{document}